%% file: main.tex
\documentclass[10pt,twocolumn,letterpaper]{article}

\usepackage[pagenumbers]{wacv} 

\usepackage{graphicx}
\usepackage{amsmath}
\usepackage{amssymb}
\usepackage{booktabs}
\usepackage{tikz}
\newcommand*\circled[1]{\tikz[baseline=(char.base)]{
            \node[shape=circle,draw,inner sep=1pt] (char) {#1};}}
\input{math_commands.tex}
\usepackage{graphbox}
\usepackage{multirow}
\usepackage{makecell}
\usepackage[accsupp]{axessibility}
\pdfoutput=1

\usepackage[pagebackref,breaklinks,colorlinks]{hyperref}

\usepackage[capitalize]{cleveref}
\crefname{section}{Sec.}{Secs.}
\Crefname{section}{Section}{Sections}
\Crefname{table}{Table}{Tables}
\crefname{table}{Tab.}{Tabs.}

\def\wacvPaperID{592} 
\def\confName{WACV}
\def\confYear{2024}

\begin{document}

\title{Expanding Expressiveness of Diffusion Models with Limited Data via Self-Distillation based Fine-Tuning}

\author{Jiwan Hur, Jaehyun Choi, Gyojin Han, Dong-Jae Lee, and Junmo Kim \\
School of Electrical Engineering, KAIST, South Korea \\
{\tt\small \{jiwan.hur, chlwogus, gjhan0820, jhtwosun, junmo.kim\}@kaist.ac.kr}
}

\maketitle

\begin{abstract}
Training diffusion models on limited datasets poses challenges in terms of limited generation capacity and expressiveness, leading to unsatisfactory results in various downstream tasks utilizing pretrained diffusion models, such as domain translation and text-guided image manipulation.
In this paper, we propose Self-Distillation for Fine-Tuning diffusion models (SDFT), a methodology to address these challenges by leveraging diverse features from diffusion models pretrained on large source datasets. 
SDFT distills more general features (shape, colors, etc.) and less domain-specific features (texture, fine details, etc) from the source model, allowing successful knowledge transfer without disturbing the training process on target datasets. 
The proposed method is not constrained by the specific architecture of the model and thus can be generally adopted to existing frameworks.
Experimental results demonstrate that SDFT enhances the expressiveness of the diffusion model with limited datasets, resulting in improved generation capabilities across various downstream tasks.
\end{abstract}

\input{sections/1_introduction}

\input{sections/2_background}
\input{sections/3_method}
\input{sections/4_experiments}

\input{sections/5_discussion}

{\small
\bibliographystyle{ieee_fullname}
\bibliography{egbib}
}

\newpage
\clearpage
\appendix

\input{supp}

\end{document}

%% file: math_commands.tex
\usepackage{amsmath,amsfonts,bm,amssymb,mathtools,amsthm}
\usepackage{color,xcolor,xspace}
\usepackage{booktabs}
\usepackage{thm-restate}

\newcommand{\norm}[1]{{\lVert {#1} \rVert}}

\def\eqref#1{Eq.~(\ref{#1})}

\def\1{\bm{1}}

\def\rmI{{\mathbf{I}}}

\def\vx{{\bm{x}}}

\DeclareMathAlphabet{\mathsfit}{\encodingdefault}{\sfdefault}{m}{sl}
\SetMathAlphabet{\mathsfit}{bold}{\encodingdefault}{\sfdefault}{bx}{n}

\def\gL{{\mathcal{L}}}

\def\gN{{\mathcal{N}}}

%% file: sections/1_introduction.tex
\section{Introduction}
\label{sec:intro}



Recently, Diffusion Models~\cite{sohl2015deep, ho2020denoising} (DMs) have emerged as a powerful family of generative models due to their diverse and high-quality image generation capability.
While generative adversarial networks (GANs)~\cite{GAN} show powerful generating capabilities in synthesizing high-quality images, they are known to have poor mode coverage~\cite{salimans2016improved, zhao2018bias}.
On the other hand, diffusion models are formulated to approximate the data distribution through likelihood estimation with a denoising score matching~\cite{song2020score}, and diffusion models trained on large datasets such as ImageNet~\cite{deng2009imagenet} outperform state-of-the-art GAN-based methods~\cite{brock2018large}, in image generation in terms of image fidelity and diversity~\cite{dhariwal2021diffusion}.

\begin{figure}[!t]
\centering
\includegraphics[width=.95\linewidth]{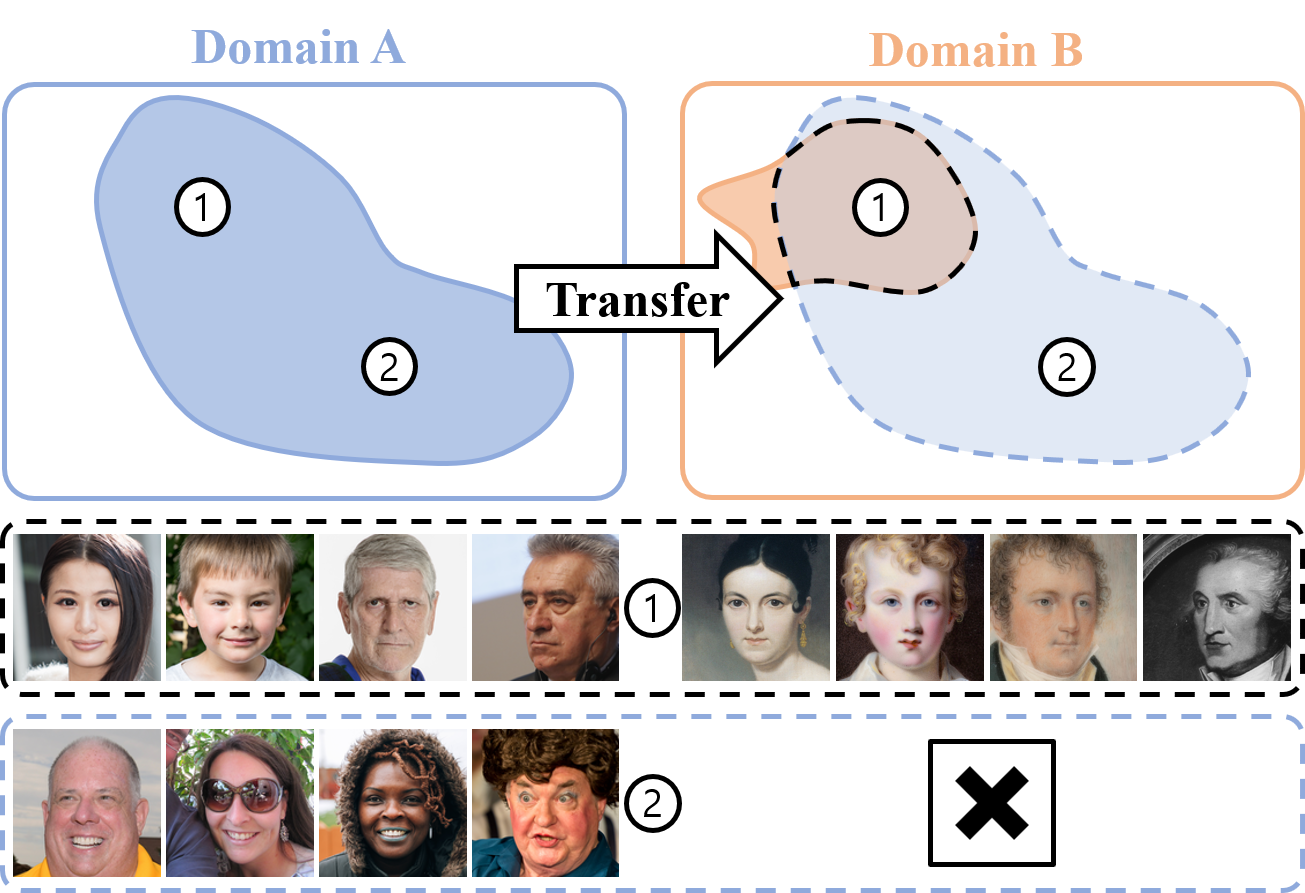}
\caption{Compared to the large datasets on Domain A (e.g. FFHQ~\cite{karras2019style}), limited datasets on domain B (e.g. MetFaces~\cite{karras2020training}) constrain the ability for diverse image generation and manipulation in Domain B. The goal of this paper is to utilize a model pre-trained on Domain A (source), and effectively transfer the diverse knowledge while training on Domain B (target).
}
\label{fig:overall}
\end{figure}

However, despite intensive research on diffusion models using large-scale datasets, there has been relatively little focus on training diffusion models on limited datasets.
Limited datasets, compared to large datasets, are more susceptible to bias and often lack diversity in terms of the range of images and attributes they contain.
For instance, MetFaces~\cite{karras2020training}, containing approximately 1K faces from medieval artworks, lacks diversity in facial attributes when compared to FFHQ~\cite{karras2019style}, which contains 70K real human faces as shown in \cref{fig:overall}.
In each domain, both datasets exhibit some shared facial attributes (region \circled{1}).
However, different from the FFHQ, MetFaces does not contain more various facial attributes (region \circled{2}) such as skin and hair colors, facial expressions, accessories, \textit{etc.}
Consequently, diffusion models trained on limited datasets may have limited \textit{expressiveness}, generating less diverse outputs and exhibiting biases in their representation.

\begin{figure}[!t]
    \centering
    \includegraphics[width=.9\linewidth]{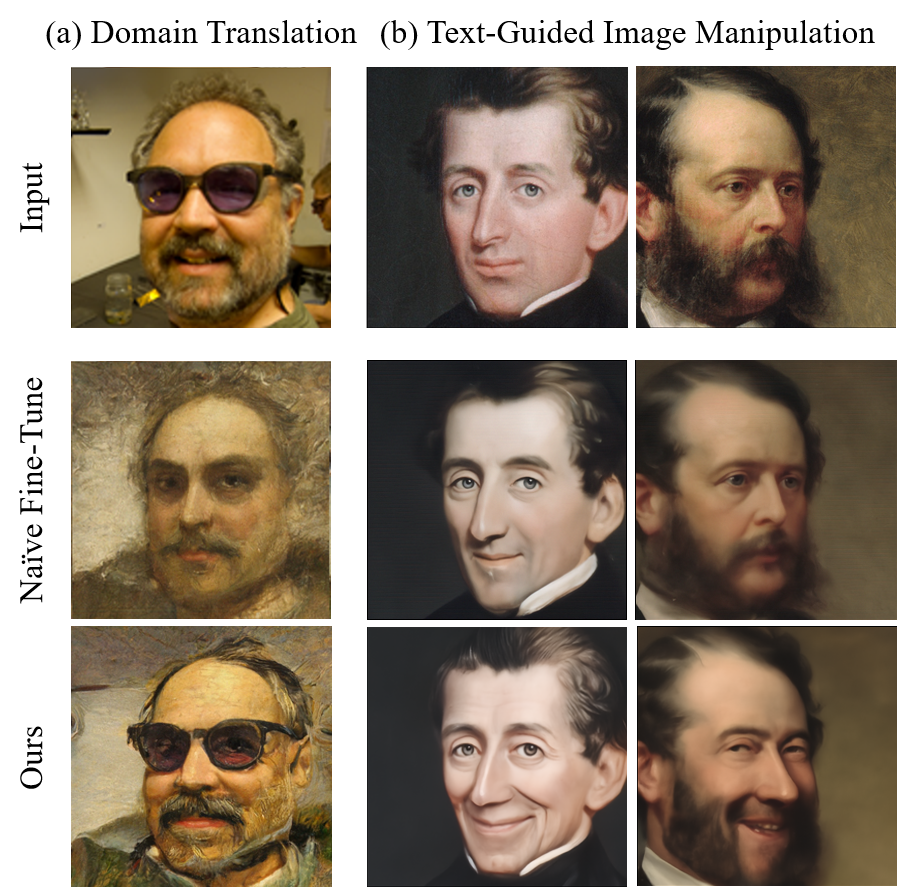}
    \caption{
    Results of downstream tasks which utilize diffusion models trained on MetFaces, such as (a) domain translation~\cite{meng2021sdedit} and (b) text-guided image manipulation~\cite{kwon2022diffusion} (script: \textit{smiling}). Since their performance is highly affected by the expressiveness of the model, na\"ively fine-tuning the model pretrained on FFHQ~\cite{karras2019style} results in the loss of crucial attributes (sunglasses) in (a) and restricted facial expression in (b) (smiling).
    With our proposed fine-tuning method, SDFT, the model can inherit diverse attributes from the source model, effectively resolving these problems.
    }
    \label{fig:intro2}
\end{figure}

The limited expressiveness of diffusion models not only hampers the generation capability of the model but also results in unsatisfactory outputs in various downstream tasks such as domain translation~\cite{zhao2022egsde,choi2021ilvr,zhao2022egsde} and text-guided image manipulation~\cite{kim2022diffusionclip,kwon2022diffusion} since these methods heavily depend on diffusion models to generate plausible outputs. 


To mitigate the aforementioned problems, in this paper, we aim to utilize diverse knowledge from the source diffusion model which is trained on large diverse datasets.
However, na\"ively fine-tuning the model on limited datasets can lead to a well-known catastrophic forgetting problem, where the model loses diverse knowledge during the fine-tuning process.
To this end, we propose Self-Distillation for Fine-Tuning diffusion models (SDFT), which leverages the diverse features from the diffusion models pretrained on large source datasets.
Specifically, to successfully transfer diverse knowledge from the source model without disturbing the training process on target datasets, SDFT prioritizes distilling general features (shape, color, \textit{etc.}) from the source model, while less emphasizing domain-specific features (texture, fine details, \textit{etc.}) from the source model.
Furthermore, we propose an auxiliary input to effectively transfer more diverse information from the source model with limited datasets.
It is worth noting that SDFT is not constrained by the specific architecture of the model, thus it can be generally adopted by existing frameworks.

Throughout various experiments, we show that the enhanced expressiveness of diffusion models by the SDFT helps to generate more diverse attributes in various downstream tasks even though they are not contained in the target limited datasets, such as sunglasses and wide smile in \cref{fig:intro2}.
Furthermore, we present that enhanced expressiveness also can be beneficial to unconditional image generation, as it helps to generate diverse and high-fidelity images.

%% file: sections/2_background.tex
\section{Background}

\subsection{Diffusion Models} \label{sec:related_diffusion}
Diffusion Models (DMs)~\cite{sohl2015deep, ho2020denoising} perturb the complex data with the tractable noise, and aim to recover the data from the noise.
Specifically, \textit{forward process} is a Markov chain that gradually perturbs the input data $\vx_0$ with Gaussian noise $\epsilon \sim \gN(0,\rmI)$ and DMs $\epsilon_\theta$ learn the inversion of the forward process, called \textit{reverse process}, where the joint distribution $p_\theta(\vx_{0:T})$ denotes the reverse process running from timestep $T$ to $0$.
Then, DMs are trained to predict noise given time \textit{t} with an objective function 
\begin{equation}    
    \gL(\epsilon_\theta) := \sum_{t=1}^{T}w_t\left[ \frac{\beta_t}{(1-\beta_t)(1-\Bar{\alpha}_t)} \norm{\epsilon_{\theta}(\vx_t,t) - \epsilon}_2^2 \right],
\end{equation}
where $\beta_t$ is a predefined noise schedule, which is a strictly decreasing function of time $t \in [0, T]$ and $\Bar{\alpha}_t:=\prod_{i=1}^{t} \alpha_i=\prod_{i=1}^{t}{1-\beta_t}$.
Ho et al.~\cite{ho2020denoising} empirically found that loss function with $w_t = (1-\beta_t)(1-\Bar{\alpha}_t)/\beta_t$ shows better results.
Recently, Choi et al.~\cite{choi2022perception} propose a weighting scheme, perception prioritized weighting (P2 weighting), which considers a signal-to-noise ratio (SNR)~\cite{kingma2021variational}, such that 
\begin{equation}    
    w'_t = \frac{w_t}{(k+\text{SNR}(t))^\gamma},
    \label{eq:p2}
\end{equation}
where SNR is strictly decreasing function of time $t$ and can be defined as $\text{SNR}(t) = \Bar{\alpha}_t/(1-\Bar{\alpha}_t)$.
P2 weighting weights more in time steps with large SNR. This allows the model to focus more on high-level context.
From the trained diffusion model, the realistic image $\vx_0$ can be sampled from the initial noise $\vx_T$ via stochastic Markovian sampling process~\cite{sohl2015deep, ho2020denoising}. However, it takes several hundred to thousand sampling steps to generate an image.
Song et al.~\cite{song2020denoising} propose denoising diffusion implicit models (DDIMs), which break the Markov chain and allow the generation of reasonable samples with few generative steps. Additionally, DDIM provides deterministic sampling from the initial noise under the specific hyperparameter setting.


\subsection{Image Translation in DMs} \label{sec:i2i}

Earlier works on image translation have been studied using GANs and shown promising results \cite{choi2018stargan,lee2023fix,yang2022pastiche,patashnik2021styleclip}.
However, they often fail to deal with various real-world images due to their limited mode coverage \cite{kim2022diffusionclip,song2022editing}.
Since this drawback prevents a range of real-world applications, DMs have received great attention for various image translation tasks.

\textbf{Domain Translation.}
The key concept behind domain translation in DMs~\cite{meng2021sdedit,choi2021ilvr,zhao2022egsde} lies in leveraging the powerful generalization ability and expressiveness of the model $\epsilon^{trg}$ which is trained solely on the target datasets.
Stochastic Differential Editing (SDEdit)~\cite{meng2021sdedit} demonstrates that by leveraging noise-perturbed data from the other domain, $\vx_t$, DMs can effectively translate $\vx_t$ to the target domain $\vx_0^{trg}$ through iterative denoising using $\epsilon^{trg}$.
Energy-Guided Stochastic Differential Equation (EGSDE)\cite{zhao2022egsde} further utilizes domain-specific and domain-independent energy functions during the sampling process for the more realistic and faithful translation, namely, domain classifier and low-pass filter, respectively.
While these approaches have shown remarkable results, their abilities to capture and translate diverse attributes totally depend on the pretrained DMs.


\textbf{Text-Guided Image Translation.}
Text-guided image translation in DMs~\cite{kim2022diffusionclip, kwon2022diffusion} mostly aims to fine-tune the pretrained DMs with a CLIP~\cite{radford2021learning}.
DiffusionCLIP~\cite{kim2022diffusionclip} proposed to fine-tune the whole diffusion model with CLIP loss for robust image editing.
Recently, the work by Kwon et al.~\cite{kwon2022diffusion}, namely, Asyrp, demonstrates that fine-tuning only the deepest layer of the U-Net, instead of the entire diffusion model, leads to a more scalable, robust, and efficient fine-tuning process.
However, since these methods fine-tune the diffusion model without further training datasets, their expressiveness also relies on the pretrained DMs.



\subsection{Fine-Tuning Unconditional Diffusion Models}
To our knowledge, fine-tuning unconditional DMs has not been widely researched yet. Recently, Moon et al.~\cite{moon2022fine} introduced a fine-tuning method to prevent overfitting during training on limited datasets. They fix the pretrained model and introduce a learnable time-aware adapter that fine-tunes the attention block of the diffusion model. 
However, as their primary objective is to prevent overfitting, they do not consider transferring diverse knowledge from the source domain to the target domain.





%% file: sections/3_method.tex
\section{Methods}

\subsection{Problem Definition}
In this paper, we consider the training of diffusion models on limited datasets.
As previously noted in the introduction and illustrated in \cref{fig:overall}, limited datasets tend to exhibit a reduced degree of diversity and inherent biases compared to large datasets.
As a remedy to this issue, we fine-tune diffusion model $\epsilon_\phi^{trg}$ which is initialized with source model $\epsilon_\theta^{src}$ pretrained on a large source dataset. 
To successfully inherit the diverse information, while avoiding catastrophic forgetting, we distill the knowledge from fixed source diffusion model $\epsilon_\theta^{src}$ during the training of the target diffusion model $\epsilon_\phi^{trg}$.
In the following sections, we provide a comprehensive explanation of the proposed distillation scheme.
To simplify the presentation of our formulas, we omit the parameters of source and target diffusion models, $\theta$ and $\phi$.

\subsection{Fine-Tuning Diffusion Models with Distillation} \label{sec:method_distill}
In this section, we describe our approach to effectively distilling diverse information from the source model $\epsilon^{src}$ to the target model $\epsilon^{trg}$, while ensuring the training on target datasets remains undisturbed.
More specifically, we desire to distill more general features (color, shape, \textit{etc}.) and less the domain-specific features (texture, fine details, \textit{etc}.) from the source model $\epsilon^{src}$.

For a similar objective, prior research on GANs has selected specific feature spaces of the generator and distilled the information in feature spaces~\cite{lee2023fix}. 
However, their applicability may be constrained by the specific model architecture, such as StyleGAN2~\cite{karras2020analyzing} and careful design and analysis on the feature space need to be preceded~\cite{karras2019style, karras2020analyzing, wu2021stylealign}.

\begin{figure}[!t]
    \centering
    \includegraphics[width=\linewidth]{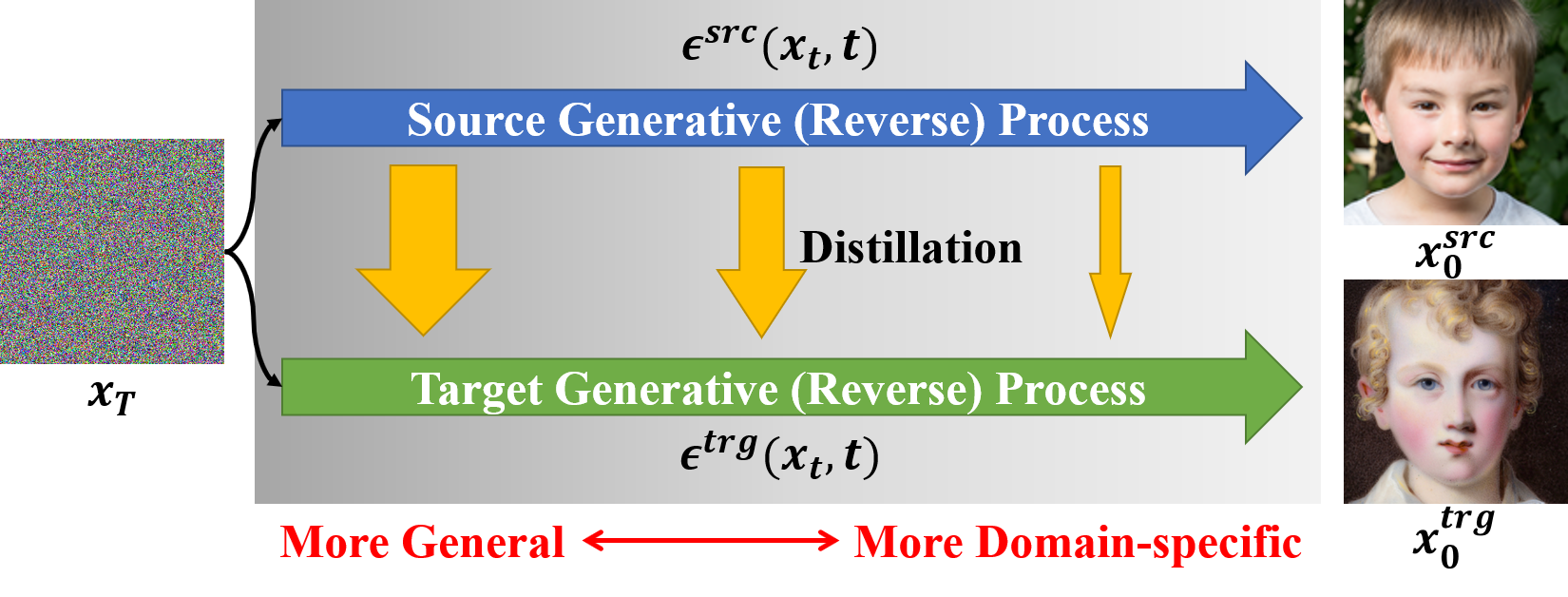}
\caption{We employ weighted distillation approach which prioritizes the general features (color, shape, pose, \textit{etc}.), while allocating lesser emphasis on domain-specific features (texture, fine-details, \textit{etc.}).
This allows the target model $\epsilon^{trg}$ to inherit various features from the source model $\epsilon^{src}$, ensuring that the training process within the target domain remains undisturbed.
}
\label{fig:method}
\end{figure}

To avoid these limitations and establish a more general methodology for DMs, we choose to distill the prediction of each time step. From the perturbed input data in target domain $\vx_t^{trg}$, we distill the knowledge by matching each prediction of source and target diffusion models:
\begin{align}
    \gL & ^{distill}(\epsilon^{trg})= \nonumber \\
    & \sum_{t=1}^{T} \left[w_{t}^{distill} \cdot \norm{\epsilon^{src}(\vx_t^{trg},t) - \epsilon^{trg}(\vx_t^{trg},t)}_2^2\right],
    \label{eq:loss_distill}
\end{align}
where $w_t^{distill}$ denotes the distillation weight on time $t$.

Before deciding the $w_t^{distill}$, we emphasize that distinct from other generative models, DMs are known to generate images by the iterative reverse process where coarse features are generated initially, and fine details are integrated later~\cite{choi2022perception}. 
In other words, during the reverse process $p_\theta(\vx_{0:T})$, DMs synthesize general features in low SNR (large $t$) and gradually synthesize more domain-specific outputs in large SNR (small $t$). 
%
%
Thus we can set $w_t^{distill}$ to inversely proportional to the SNR($t$) 
\begin{equation}    
    w^{distill}_t = \frac{w_t}{(k+\text{SNR}(t))^{\gamma^{distill}}},
    \label{eq:w_distill}
\end{equation}
which has a same formulation with P2 weighting in \cref{eq:p2}. 
However, we note that while P2 weighting aims to help diffusion models focus more on perceptually rich contents of the datasets, our proposed distillation weighting scheme is designed to distill more general features from the source model to the target model.
With $w_t^{distill}$, the target model $\epsilon^{trg}$ can preserve general diverse features from source model $\epsilon^{src}$. Moreover, it can be universally applied to any diffusion model, as it does not rely on the specific architecture of the model. \cref{fig:method} shows the overall illustration of the proposed distillation method.

However, while the distillation loss is beneficial in preserving the diversity from the source model, there still remain several considerations to be addressed. In the following sections, we tackle several expected problems and provide solutions if necessary.

\subsection{Does \texorpdfstring{$\epsilon^{src}(\vx_t^{trg},t)$}{} Generate Meaningful Output?}\label{sec:ood}
In \cref{eq:loss_distill}, the source model $\epsilon^{src}$ generates output from the input $\vx_t^{trg}$, which is a target domain sample perturbed by the Gaussian noise.
Since $\epsilon^{src}$ is only trained with source datasets, $\vx_t^{trg}$ can be considered as out-of-domain data, which may result in the failure to generate meaningful outputs from $\epsilon^{src}$, leading to unsuccessful distillation.

Insights into this issue can be gained from previous research. Deja et al.~\cite{deja2022analyzing} found that DMs have generalizability on other data distributions. 
Furthermore, studies on domain translation methods such as SDEdit~\cite{meng2021sdedit} described in \cref{sec:i2i} demonstrated that adding noise to images from a similar but distinct domain and then running the reverse diffusion process can yield reasonable in-domain outputs.
Thus, even with out-of-domain inputs, diffusion models can effectively convert into meaningful in-domain outputs, which enables successful distillation using \cref{eq:loss_distill}.
%
%
Moreover, SDEdit~\cite{meng2021sdedit} demonstrates that running the reverse process in small $t$ only changes domain-specific features (e.g. texture) from the input image while maintaining general features (e.g. silhouette). 
This means that the reverse process of DMs in small $t$ generates domain-specific features while that in large $t$ generates general features, which gives more justification for using $w^{distill}_t$ to prioritize general features over domain-specific features in \cref{eq:w_distill}.


\begin{figure}[!t]
    \centering
    \includegraphics[width=.7\linewidth]{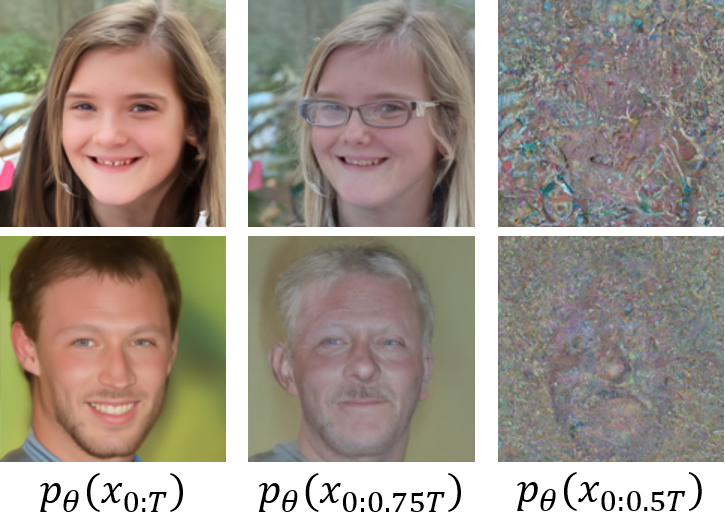}
\caption{Images in each row are sampled from the same initial noises $p(\vx_{T'})=\gN(0,\rmI)$, but they take different \textit{partial reverse process} $p_\theta(\vx_{0:T'})$. It shows that when the $T'$ is large, diffusion models can generate reasonable outputs from pure noise.}
\label{fig:agn}
\end{figure}

\subsection{Distilling More Diverse Features} \label{sec:agn}
Expanding on the issues outlined in \cref{sec:ood}, using $\vx_t^{trg}$ as input introduces an additional unresolved problem.
Remember that we consider the target datasets which include limited samples with reduced diversity compared to the source datasets.
As a result, we can not fully extract the diverse features inherent within $\epsilon^{src}$ using limited input $\vx_t^{trg}$.

To address this issue, we propose to use auxiliary inputs to distill more diverse features from $\epsilon^{src}$, without accessing the source or additional datasets.
Using proxy inputs to transfer diverse knowledge from the teacher network has been widely explored in data-free knowledge distillation~\cite{yin2020dreaming,zhang2021data}. 
They demonstrated that the utilization of synthesized input data effectively facilitates the transfer of diverse knowledge from the teacher network, even though these data significantly differ from the original source data.
Inspired by this, we propose an auxiliary input specifically designed for the diffusion models to distill more diverse features from $\epsilon^{src}$.
Notably, we discovered that diffusion models possess the capability to generate plausible outputs from the pure noise $\vx_T$, without requiring a perturbed image $\vx_t$, in the initial reverse process.
That means, diffusion models can generate meaningful outputs from pure noise in the initial reverse process.
\cref{fig:agn} shows sampled outputs obtained from the same initial noise in each row, but applying various \textit{partial reverse process} $p_\theta(\vx_{0:T'}),$ defined in \cref{sec:related_diffusion}($(T'<T)$).
In the second row, even though the initial reverse process ranging from T to 0.75T is skipped, diffusion models can generate reasonable outputs, albeit with some color degradation.
To this end, we propose additional loss $\gL^{aux}$ that utilizes pure noise as an auxiliary input for a diverse feature distillation:
\begin{align}
    \gL  ^{aux}(\epsilon^{trg})= \sum_{t=1}^{T} \left[w_{t}^{aux} \cdot \norm{\epsilon^{src}(\vx_T,t) - \epsilon^{trg}(\vx_T,t)}_2^2\right],
    \label{eq:aux}
\end{align}
where $w_{t}^{aux}$ is a same weighted distillation scheme in \cref{eq:w_distill}, but using different hyperparameter $\gamma^{aux}$. 
As depicted in the third column of \cref{fig:agn}, since the output drastically collapses as $T'$ decreases, we set high $\gamma^{aux}$ to make $w_{t}^{aux}$ nearly 0 for a small $t$.



\subsection{Total Loss Function with SDFT}
To summarize, the proposed fine-tuning method with distillation uses the loss function:
\begin{equation}
    \gL^{total}= \gL^{diffusion} + \lambda^{distill}\gL^{distill} + \lambda^{aux}\gL^{aux},
\end{equation}
where $\gL^{diffusion}$ is a objective function of base diffusion model and $\lambda^{distill}$ and $\lambda^{aux}$ is a hyperparameter. 

%% file: sections/4_experiments.tex
\section{Experiments}

\begin{figure*}[!t]
    \centering
    \includegraphics[width=.9\textwidth]{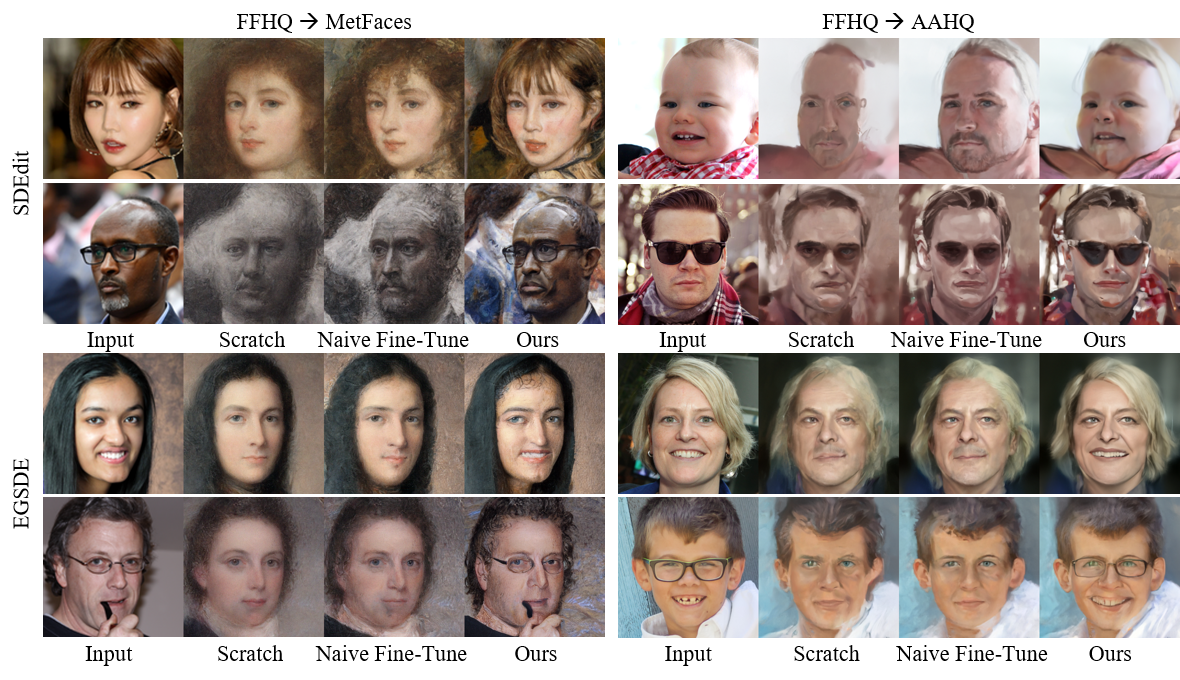}
\caption{The generated images sampled utilizing the domain translation method denoted above.}
\label{fig:qual_i2i}
\end{figure*}

\begin{table*}[!h]
\centering
\begin{tabular}{c|c|c||ccccc}

 Dataset & Translation Method & Training Method & PSNR $\uparrow$ & SSIM $\uparrow$ &  LPIPS $\downarrow$ & FID $\downarrow$ & KID (\footnotesize{$\times 10^3$}) $\downarrow$\\
 \hline\hline
 {\multirow{6}{*}{\shortstack{MetFaces \\\cite{karras2020training}}}} & \
 {\multirow{3}{*}{SDEdit~\cite{zhao2022egsde}}}   & Scratch      & 13.8 & 0.31 & 0.579 & 74.63 & 52.45   \\
                            && Na\"ive Fine-Tune & 14.95 & 0.309 & 0.533 & \textbf{56.42} & \textbf{37.00}   \\
                            && Ours (SDFT)                       & \textbf{16.44} & \textbf{0.353} & \textbf{0.481} & 65.18 & 40.95   \\ \cline{2-8}
 & {\multirow{3}{*}{EGSDE~\cite{zhao2022egsde}}}    & Scratch      & 15.15 & 0.31 & 0.539 & 89.67 & 71.14  \\
                            && Na\"ive Fine-Tune & 16.03 & 0.333 & 0.509 & \textbf{70.02} & 50.52  \\
                            && Ours (SDFT)                       & \textbf{17.42} & \textbf{0.392} & \textbf{0.440} & 70.78 & \textbf{43.84}  \\ \hline
 {\multirow{6}{*}{\shortstack{AAHQ}}} & \
 {\multirow{3}{*}{SDEdit~\cite{zhao2022egsde}}}   & Scratch      & 14.26 & 0.362 & 0.521 & 65.54 & 63.42   \\
                            && Na\"ive Fine-Tune & \textbf{14.60} & 0.374 & 0.489 & 54.75 & 48.69   \\
                            && Ours (SDFT)                       & 14.59 & \textbf{0.369} & \textbf{0.486} & \textbf{51.12} & \textbf{44.46}   \\  \cline{2-8}
 & {\multirow{3}{*}{EGSDE~\cite{zhao2022egsde}}}    & Scratch      & 15.91 & 0.407  & 0.464  & 82.00 & 78.26  \\
                            && Na\"ive Fine-Tune & \textbf{16.18} & 0.422  & 0.430  & 65.14 & 57.30   \\
                            && Ours (SDFT)                       & 16.13 & \textbf{0.421} & \textbf{0.423} & \textbf{60.49} & \textbf{52.24}   \\ 
\end{tabular}
\caption{Quantitative results of domain translation methods using diffusion models trained with various methods. }
\label{tab:quant}
\vspace{-1.5mm}
\end{table*}

\subsection{Experimental Setup}
In this section, we verify the effectiveness of SDFT for fine-tuning DMs to inherit expressiveness from the source models when the target datasets have limited samples and attributes.
We present that the enhanced expressiveness of DMs results in improvements in various downstream image translation tasks including domain translation and text-guided image manipulation.
To our knowledge, as the open source implementation of the fine-tuning method for unconditional diffusion model~\cite{moon2022fine} is not publicly available, we compare SDFT with a model trained from scratch and na\"ively fine-tuned model, referred to as the Scratch model and Na\"ive Fine-Tune model, respectively. For simplicity, we also name the model trained with SDFT as SDFT.

\textbf{Datasets.}
We use FFHQ~\cite{karras2019style} for the source dataset, which has 70K real faces with various attributes.
For the target limited datasets, we utilize MetFaces\cite{karras2020training}, which has 1,336 high-quality portraits. Due to the limited samples and inherent biases, MetFaces do not or scarcely contain diverse facial attributes (e.g. smiling with teeth, sunglasses, various hairstyles \textit{etc.}). 
We further utilize the AAHQ\cite{liu2021blendgan} for the target dataset, which has 25K high-quality artistic faces. For the limited dataset, we select images from AAHQ using CLIP following the nie et al.~\cite{nie2021controllable}. As a result, we utilize 1,437 images that contain expressionless males without glasses.
All images are resized to 256$\times$256 resolution. 
A detailed explanation for preparing the dataset is provided in the supplementary material.

\textbf{Implementation detail.}
For all experiments, we utilize officially implemented ADM architecture~\cite{dhariwal2021diffusion} and public checkpoint pretrained on FFHQ for the source model~\cite{choi2022perception}.
For efficiency, we use 40-step deterministic DDIM sampling~\cite{song2020denoising} for all experiments. 
We train all models until the 80k training iterations and report the best results.
The hyperparameters can be found in supplementary material.

\begin{figure*}[!h]
    \centering
    \includegraphics[width=\textwidth]{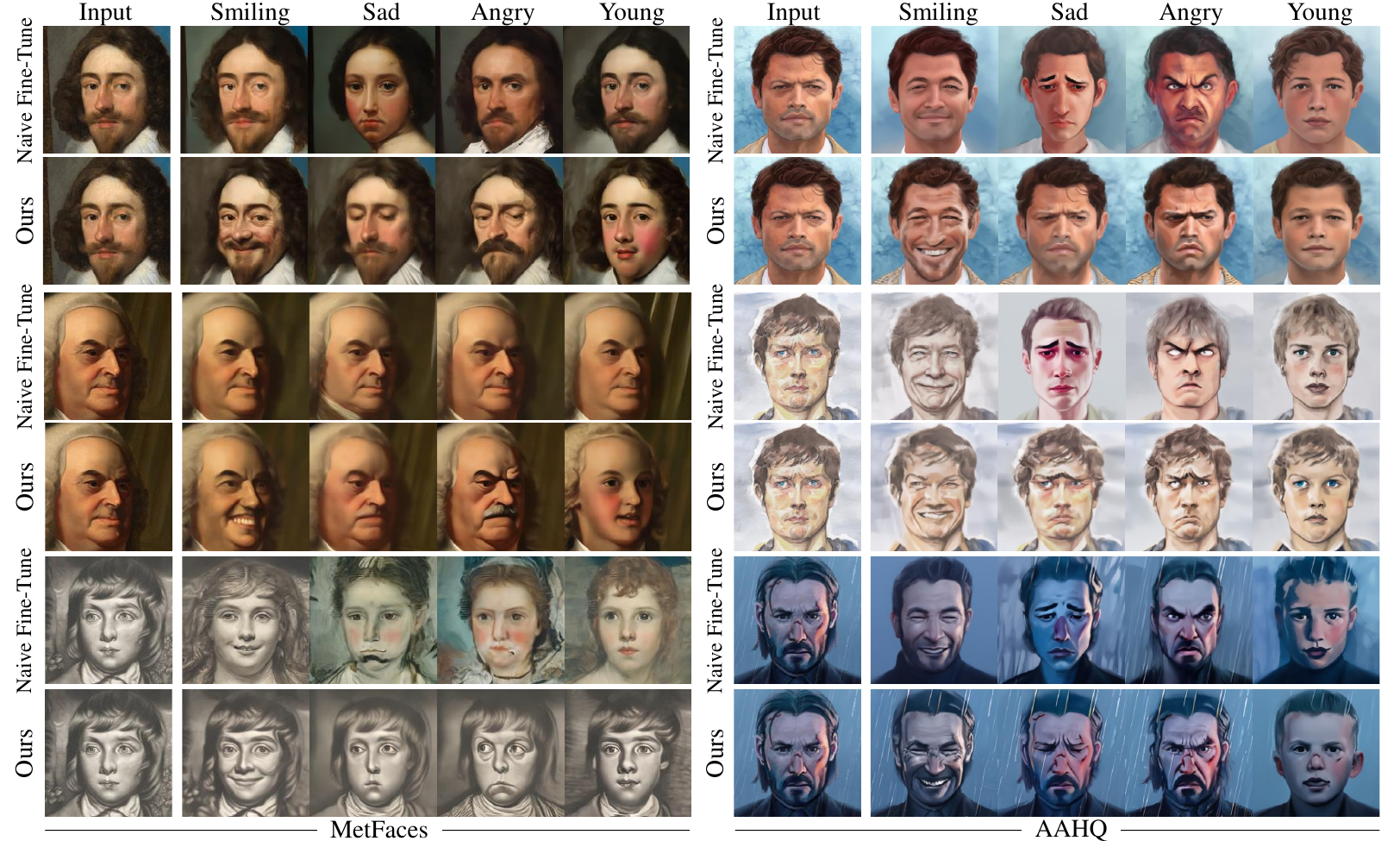}
\caption{Results of Asyrp~\cite{kwon2022diffusion} from Na\"ive Fine-Tune model and Ours (SDFT). SDFT can express more diverse facial attributes. }
\label{fig:text_guided}
\vspace{-1mm}
\end{figure*}

\subsection{Results in Domain Translation}
We present that the enhanced expressiveness of diffusion models can greatly improve the performance of domain translation tasks.
For successful domain translation, the translated image should be \textit{realistic} to fit the style of the target domain and \textit{faithful} to ensure that the various attributes from the input image are accurately preserved.

\textbf{Qualitative Comparison.}
\cref{fig:qual_i2i} shows the qualitative comparison between scratch, na\"ive fine-tune, and ours (SDFT) with domain translation methods, SDEdit~\cite{zhao2022egsde} and EGSDE~\cite{zhao2022egsde}.
Since these methods utilize diffusion models trained on the target domain, the expressiveness of the model determines the success of the translation.
Domain translation outputs with the scratch and Na\"ive fine-Tune model show less \textit{faithful} outputs due to the limited training datasets, failing to translate \textit{unseen} attributes from the training data such as glasses, and facial expressions. Furthermore, the translated outputs show biased representations such as translating baby and female to male in 3rd row.
With a proposed fine-tuning approach, SDFT resolves the above problems and shows more \textit{realistic} translated outputs, while showing reasonable \textit{faithful}.

\textbf{Quantitative Comparison.}
To measure the \textit{faithfulness}, we evaluate the similarity between input-output pairs using the peak signal-to-noise ratio (PSNR), structural similarity index (SSIM), and learned perceptual image patch similarity (LPIPS)~\cite{zhang2018unreasonable}. We randomly select 10K images from FFHQ and generate 10K translated outputs. 
For the \textit{realism}, we report the widely used Fr\'echet inception distance (FID)~\cite{heusel2017gans} and kernel inception distance (KID)~\cite{binkowski2018demystifying} where the latter is known to be unbiased, thus more proper to the limited datasets.
The metrics for \textit{faithful} are calculated using translated outputs and paired FFHQ inputs and metrics for \textit{realism} are calculated using translated outputs and entire target datasets. For experiments on AAHQ, we use entire datasets to calculate FID and KID.

\cref{tab:quant} shows the quantitative results. 
In each domain translation method, SDFT shows the most \textit{faithful} results, achieving the highest PSNR and SSIM, and the lowest LPIPS, except for PSNR in AAHQ. 
However, SDFT is reported as less \textit{realistic} than Na\"ive Fine-Tune model in SDEdit of MetFaces since SDFT translates attributes that are not contained in the MetFaces, as depicted in \cref{fig:qual_i2i}. 
However, using EGSDE~\cite{zhao2022egsde}, SDFT achieves the lowest KID, showing the most \textit{realistic} outputs since the domain-specific energy function, a classifier between source and target domain, naturally drives the translated outputs to the target domain, without harming the \textit{realism}. 
In limited AAHQ in which we calculate FID and KID using the entire AAHQ dataset, SDFT shows the most \textit{realistic}, indicating that SDFT can generate high fidelity and diversity samples in the target domain with limited target datasets.

\subsection{Results in Text-Guided Image Translation} \label{sec:text_guided}
We further show that the improved expressiveness of the diffusion model by the SDFT can be helpful in the more complex downstream tasks, such as text-guided image translation.
For the text-guided image translation method, we choose Asyrp~\cite{kwon2022diffusion} since it fully utilizes the expressiveness embedded in the pretrained diffusion model.
It only trains a small module that translates the deepest layer of U-net, while keeping all parameters unchanged during the fine-tuning with CLIP~\cite{radford2021learning}.
Please note that we introduce a fine-tuning method, SDFT, for the unconditional diffusion model on limited datasets and Asyrp is a fine-tuning method for the text-guided image translation using a trained unconditional diffusion model.


\textbf{Qualitative Comparison.}
\cref{fig:text_guided} shows the input images and translated images with various text guidance.
Since na\"ive fine-tuning on the MetFaces have a limited expressiveness due to the limited and biased samples, it can not generate diverse facial attributes and often fails to maintain identities.
Contrarily, SDFT can express more diverse facial attributes, while successfully preserving the identity of the face.
Even though the training datasets do not have diverse facial attributes, SDFT can express through enhanced expressiveness which is inherited from the source model.
Notably, even though we transfer the knowledge from the FFHQ, the enhanced expressiveness affects not only the outputs close to the source dataset but also the more distant outputs, such as monochrome paintings.

\begin{table}[!t]
\centering
\begin{tabular}{ccccccc} \Xhline{2\arrayrulewidth}
&& \multicolumn{2}{c}{MetFaces} && \multicolumn{2}{c}{AAHQ} \\ \cline{3-4} \cline{6-7}
&& $\mathbf{\mathcal{S}_\text{dir}} \uparrow$ & ID$\uparrow$ && $\mathbf{\mathcal{S}_\text{dir}} \uparrow$ & ID$\uparrow$ \\ \hline
Na\"ive Fine-Tune &&  0.060  & 0.760 && 0.133 & 0.318 \\
Ours (SDFT) & & \textbf{0.081} & \textbf{0.765} && \textbf{0.143} & \textbf{0.452} \\ \Xhline{2\arrayrulewidth}
\end{tabular}
\caption{Quantitative results on text-guided image manipulation.}
\label{tab:quant_text_guided}
\vspace{-1mm}
\end{table}

\textbf{Quantitative Comparison.}
We evaluate the successful text-guided manipulation in two metrics: Directional CLIP similarity ($\mathbf{\mathcal{S}_\text{dir}}$)~\cite{kim2022diffusionclip} and face identity similarity (ID). $\mathbf{\mathcal{S}_\text{dir}}$ measure the successful manipulation of input image given text guidance using pretrained CLIP~\cite{radford2021learning} embedding and ID measures the preservation of identity using pretrained face recognition models~\cite{deng2019arcface}. 
\cref{tab:quant_text_guided} shows the results using 5 text prompts (\textit{smiling, sad, angry, young and old}). SDFT outperforms the na\"ive fine-tune model, demonstrating superior semantic manipulation capabilities given text while preserving various identities.

\begin{figure}[!t]
    \centering
    \includegraphics[width=\linewidth]{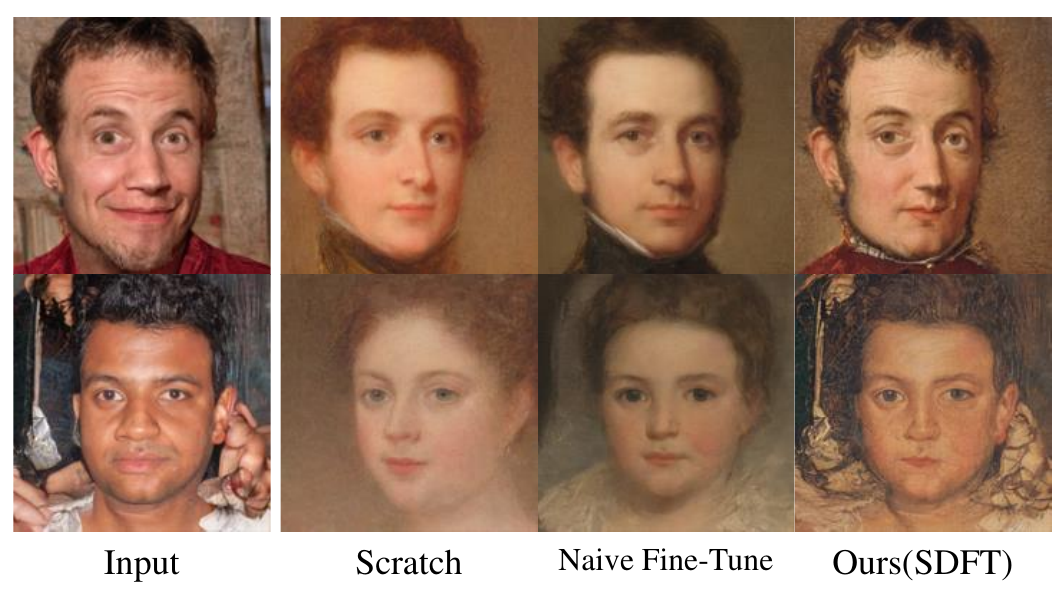}
\caption{Generated samples from unconditional image generation. Images in each row are generated from the same initial noise.  }
\label{fig:uncond}
\vspace{-1mm}
\end{figure}

\subsection{Effect on Unconditional Image Generation}
Finally, we present that the benefits of SDFT are also helpful for generating diverse and high-fidelity images.
By expanding the dual diffusion implicit bridges (DDIBs)~\cite{su2022dual} theorem, the DDIM sampling between the source model and target model generates semantically aligned outputs from the same initial noise. 
A more comprehensive analysis of DDIBs and their connection to unconditional generation is provided in the supplementary material.
\cref{fig:uncond} shows the results of unconditional generation from the source model trained on FFHQ and from the various models fine-tuned on MetFaces. 
Scratch and Na\"ive Fine-Tune model shows semantically aligned images, but they fall short in generating a range of diverse attributes.
However, SDFT can generate more semantically aligned images, preserving more diverse attributes that are not included in the target datasets such as the smiling face (1st row) and people with dark skin (2nd row).
\cref{tab:quant_uncond} shows the measured FID and KID using 10K generated samples with MetFaces and the entire AAHQ dataset.
The perceptual distance (LPIPS) is measured between the 2K generated samples between the source and target model from the same initial noises.
From the results, the proposed SDFT shows the capability to generate diverse and high-fidelity images from limited datasets while preserving the diversities from the source model.

\begin{table}[!t]
\centering
\begin{tabular}{ccccc} \Xhline{2\arrayrulewidth}
& FID$\downarrow$& KID (\footnotesize{$\times 10^3$}) $\downarrow$ & LPIPS$\downarrow$ \\ \cline{2-4}
& \multicolumn{3}{c}{MetFaces} \\ \cline{2-4}
Scratch & 65.91& 44.79 & 0.488 \\
Na\"ive Fine-Tune & 43.45 & 22.81 & 0.474 \\
Ours (SDFT) & \textbf{35.11} & \textbf{17.14} & \textbf{0.46} \\ \hline
& \multicolumn{3}{c}{AAHQ} \\ \cline{2-4}
Scratch & 62.48 & 48.67 & 0.575 \\
Na\"ive Fine-Tune & 64.08 & 56.70 & 0.562\\
Ours (SDFT) & \textbf{42.54} & \textbf{33.75} & \textbf{0.469} \\ \Xhline{2\arrayrulewidth}
\end{tabular}
\caption{Quantitative results on unconditional image generation.}
\label{tab:quant_uncond}
\end{table}

\begin{figure}[!t]
    \centering
    \includegraphics[width=.75\linewidth]{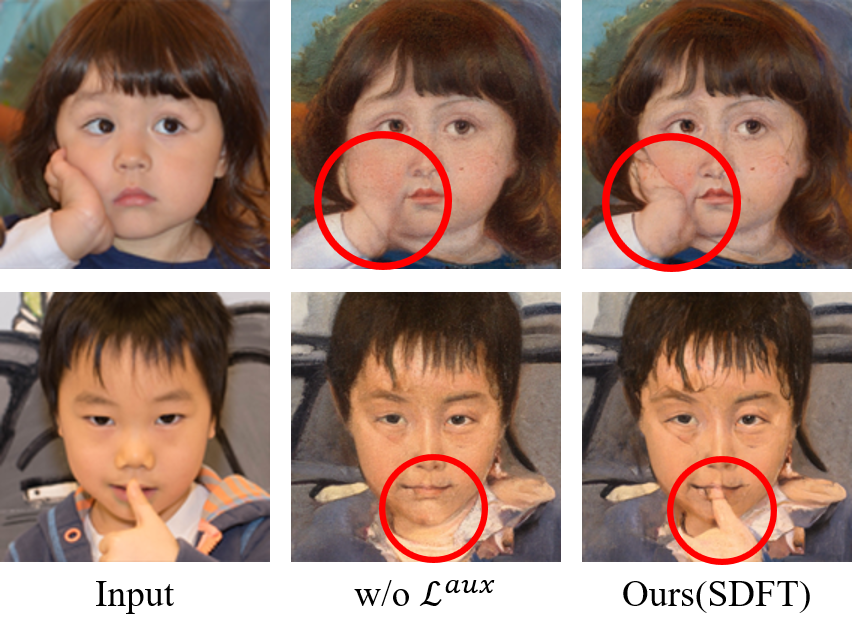}
\caption{ $\gL^{aux}$ helps DMs deal with more diverse inputs.}
\label{fig:abla}
\vspace{-1mm}
\end{figure}

\subsection{Ablation Study}
In this section, we present the effectiveness of the proposed methods using domain translation from face to portrait using MetFaces\cite{karras2020training} and EGSDE\cite{zhao2022egsde}.

\noindent \textbf{Auxiliary Input}
We present the effectiveness of the auxiliary inputs for distilling more diverse features described in \cref{sec:agn}. \cref{fig:abla} illustrates the results of domain translation on out-of-domain inputs. 
By utilizing the auxiliary inputs during the training, the more diverse features that are not included in the target datasets can be transferred and help the model deal with out-of-domain samples successfully.

\begin{figure}[!t]
    \centering
    \includegraphics[width=\linewidth]{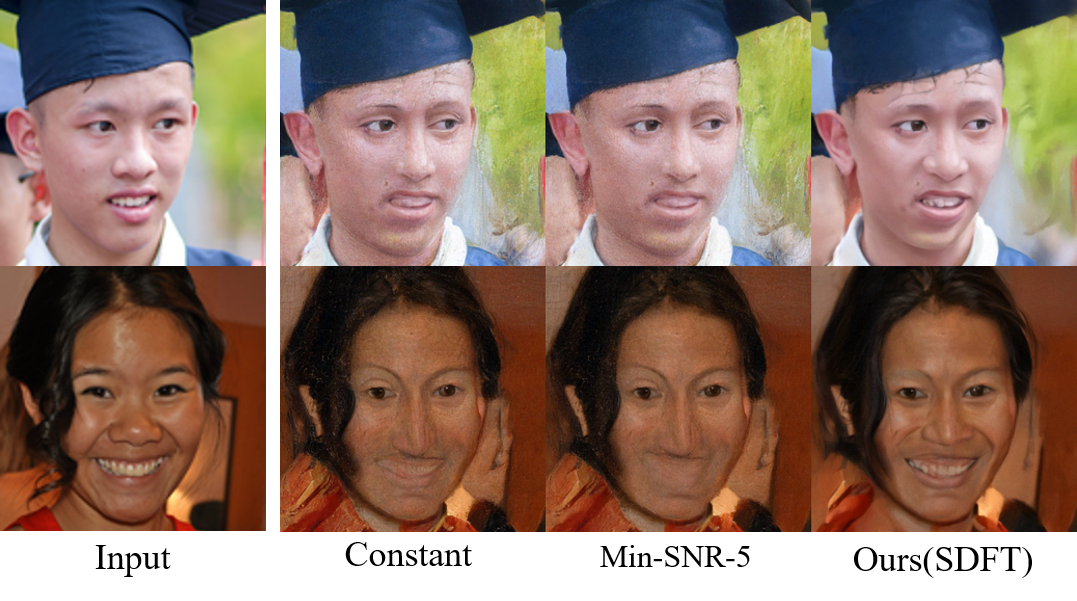}
\caption{
Effect of various distillation weights for fine-tuning.
}
\label{fig:ablation}
\vspace{-1mm}
\end{figure}

\noindent \textbf{Different Weighting Scheme}
We also compare the proposed weighting scheme for distillation with other weightings which have been used for the training of DMs such as constant weighting ($w_t^{distill}=1\cdot w_t$)\cite{ho2020denoising} and a Min-SNR weighting ($w_t^{distill}=$min\{SNR($t$),$\gamma$\} $\cdot w_t$) where $w_t$ is defined in \cref{sec:related_diffusion} and $\gamma$ is set to 5 following Hang et al.\cite{hang2023efficient}.
%
\cref{fig:ablation} shows that other methods fail to transfer attributes from source images such as teeth. However, the proposed weighting scheme prioritizes the general features while discarding domain-specific features from the source model, leading to successful domain translation. For unconditional generation in MetFaces, \cref{tab:ablation} shows that SDFT can generate more realistic (low FID and KID) images while preserving more semantics from the source model (low LPIPS) compared to other weighting strategies.
\begin{table}[!t]
\centering
\begin{tabular}{ccccc} \Xhline{2\arrayrulewidth}
& FID$\downarrow$& KID (\footnotesize{$\times 10^3$}) $\downarrow$ & LPIPS$\downarrow$ \\ \hline
Constant & 45.21 & 28.42 & 0.494  \\
Min-SNR-5 & 44.75 & 27.96 & 0.476  \\
Ours (SDFT) & \textbf{35.11} & \textbf{17.14} & \textbf{0.46} \\ \Xhline{2\arrayrulewidth}
\end{tabular}
\caption{Ablation studies on different weighting strategies.}
\label{tab:ablation}
\end{table}

%% file: sections/5_discussion.tex
\section{Conclusion and Future Works}
In this paper, we propose a self-distillation-based fine-tuning method for training diffusion models in limited datasets by leveraging the diverse knowledge from the source model trained on large datasets. 
Experimental results demonstrate that SDFT can effectively enhance the expressiveness of the diffusion models, leading to improved performance in various downstream tasks. 
%
\textbf{Future works.} 
Recently, rather than training full parameters, parameter-efficient fine-tuning (PEFT) can bring efficient and promising results \cite{ryu2023low,sohn2023styledrop}. Since the SDFT can be orthogonally combined with these methods, we leave it for future work to investigate the advantage of combining SDFT with PEFT.
%
%
Lastly, we only consider the diffusion models as noise predictors, but recent studies on distilling diffusion models found that utilizing velocity can bring effective knowledge transfer\cite{salimans2022progressive,li2023snapfusion}. Combining SDFT with various diffusion parametrizations is also an interesting future work.

\textbf{Acknowledgements.}
This work was partly supported by Institute of Information \& communications Technology Planning \& Evaluation (IITP) grant funded by the Korea government(MSIT) (No.2021-0-02068, Artificial Intelligence Innovation Hub) and the Engineering Research Center Program through the National Research Foundation of Korea (NRF) funded by the Korean Government MSIT (NRF-2018R1A5A1059921)

%% file: supp.tex
\section{Connection between DDIBs and Unconditional Image Generation} \label{sec:supp_ddib}
In this section, we provide a general analysis of Dual Diffusion Implicit Bridges (DDIBs)~\cite{su2022dual} and their connection to unconditional image generation. 
We first note that with a proper hyperparameter, a particular sampling process of diffusion model, denoising diffusion implicit model (DDIM) is an ordinary differential equation (ODE) process where the output is deterministically provided given the input noise.
Leveraging the DDIM and trained diffusion models, the noise can be translated into the image through the \textit{reverse process} of diffusion models $p_\theta(\vx_{t-1}|\vx_{t})$, where $\vx_t$ is a perturbed image $\vx_0$ with a time step $t$. 
Notably, DDIM can run in the forward direction $p_\theta(\vx_t|\vx_{t-1}$) to get the noise from the image.
Since $p_\theta(\vx_t|\vx_{t-1})$ is also deterministic in DDIM, the sampled noise can be used as a latent of the image.

Recently, Su et al.~\cite{su2022dual} propose an unpaired domain translation method, Dual Diffusion Implicit Bridges (DDIBs), which leverages the source diffusion models $\epsilon^{src}$ and target diffusion models $\epsilon^{trg}$ to get aligned image pairs from source and target domain.
The source model $\epsilon^{src}$ runs DDIM sampling in the forward direction to get the latent noise $\vx_T^{src}$ from the input image of the source domain $\vx_0^{src}$.
Then, from the $\vx_0^{src}$, the target model $\epsilon^{trg}$ runs DDIM sampling in the reverse direction to get an image $\vx_0^{trg}$ from the $\vx_T^{src}$.
Through the above processes, DDIBs prove that $\vx_0^{trg}$ is semantically aligned with the $\vx_0^{src}$, without requiring a joint training of source and target pairs.

From the DDIBs, it can be derived that given the same initial noise $\vx_T$, the source model $\epsilon^{src}$ and the target model $\epsilon^{trg}$ can generate aligned images, $\vx_0^{src}$ and $\vx_0^{trg}$, by running the deterministic DDIM sampling process in the reverse direction.
As a result, unconditional generation from the same initial noise can be interpreted as a domain translation between the source and target model, generating semantically aligned images.
From this point of view, the proposed method, Self-Distillation-based Fine-Tuning (SDFT), can generate a more diverse and more aligned image with the source model than the Na\"ive Fine-Tune model, as shown in the main manuscript. 

\begin{figure}[!t]
    \centering
    \includegraphics[width=\linewidth]{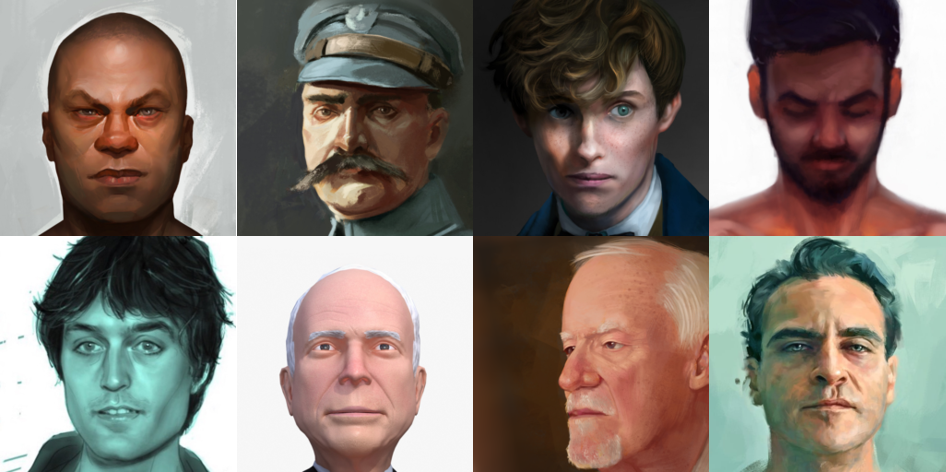}
\caption{Randomly selected images from a limited AAHQ dataset for the training of diffusion models in the main manuscript.}
\label{fig:aahq_dataset}
\end{figure}

\section{Explanation for Preparing Dataset}
We utilize source diffusion models pretrained on FFHQ, which has 70K diverse real faces with various attributes. For the target limited datasets, we utilize MetFaces\cite{karras2020training}, which has 1,336 high-quality portraits. Due to the limited samples and inherent biases, MetFaces do not or scarcely contain diverse facial attributes (e.g. smiling with teeth, sunglasses, various hairstyles \textit{etc.}). We further exclude 10 samples which include \textit{glasses} from the MetFaces for the more challenging scenario. 
For another target dataset using the same source dataset, we utilize AAHQ\cite{liu2021blendgan} which contains 25k high-quality artistic faces. However, AAHQ is a sufficiently large and diverse dataset. To simulate the limited, biased dataset, we select images from AAHQ using CLIP\cite{radford2021learning} following the nie et al.\cite{nie2021controllable}. Specifically, we measure the cosine similarity of embedding vectors of CLIP between AAHQ images and the prompt \textit{``A realistic painting of an expressionless man without glasses ''}. Then by thresholding the similarity and manually excluding some misclassified images, we get 1,437 limited AAHQ images. As described in nie et al.\cite{nie2021controllable}, using CLIP similarity for preparing datasets is efficient and effective for selecting datasets based on the natural language without labor-intensive work. 
\cref{fig:aahq_dataset} shows the randomly selected images from a limited AAHQ dataset. Note that even though it contains a similar number of images as MetFaces, it contains more various images such as various skin colors, and various emotions. For example, MetFaces contains the faces of medieval works of art, so most samples consistently show similar styles of painting and the style of painting of the time, such as smiling a little in most samples. However, despite using CLIP to select a limited and biased dataset in AAHQ, it lacks biases such as picture style and expressions.


\section{Training Details}
\begin{figure}[!h]
    \centering
    \includegraphics[width=.8\linewidth]{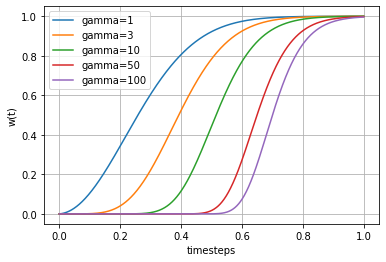}
\caption{Different weightings according to hyperparameter $\gamma$.}
\label{fig:gamma}
\end{figure}

We provide additional details for training diffusion models in the main manuscript. 
We use the lighter version of ADM \cite{dhariwal2021diffusion} for the baseline model in all experiments and use default settings. To implement the SDFT, we use 4 hyperparameters that define the 
in \cref{tab:comp_hyperparameters}. Note that since the output from the auxiliary input drastically collapses as the timestep increases, we set higher $\gamma^{aux}$ for all experiments. The visualization of $w(t)$ used for SDFT according to different gamma values is provided in \cref{fig:gamma}.

\begin{table}[h]
\centering
\scalebox{0.9}
{
\begin{tabular}{lccc}
\toprule
Datasets & MetFaces  & AAHQ \\
\midrule
     $\lambda^{distill}$ & 0.1 & 0.1\\
     $\lambda^{aux}$       & 0.1 &   0.3  \\
     $\gamma^{distill}$        & 3    & 50   \\
     $\gamma^{aux}$        & 3    & 50    \\
 \bottomrule
\end{tabular}}
\vspace{-8pt}
\caption{
 Hyperparameters for SDFT training.
\label{tab:comp_hyperparameters}}
\end{table}

\section{Experimental Details in Domain Translation}
\begin{figure*}[!ht]
    \centering
    \includegraphics[width=\textwidth]{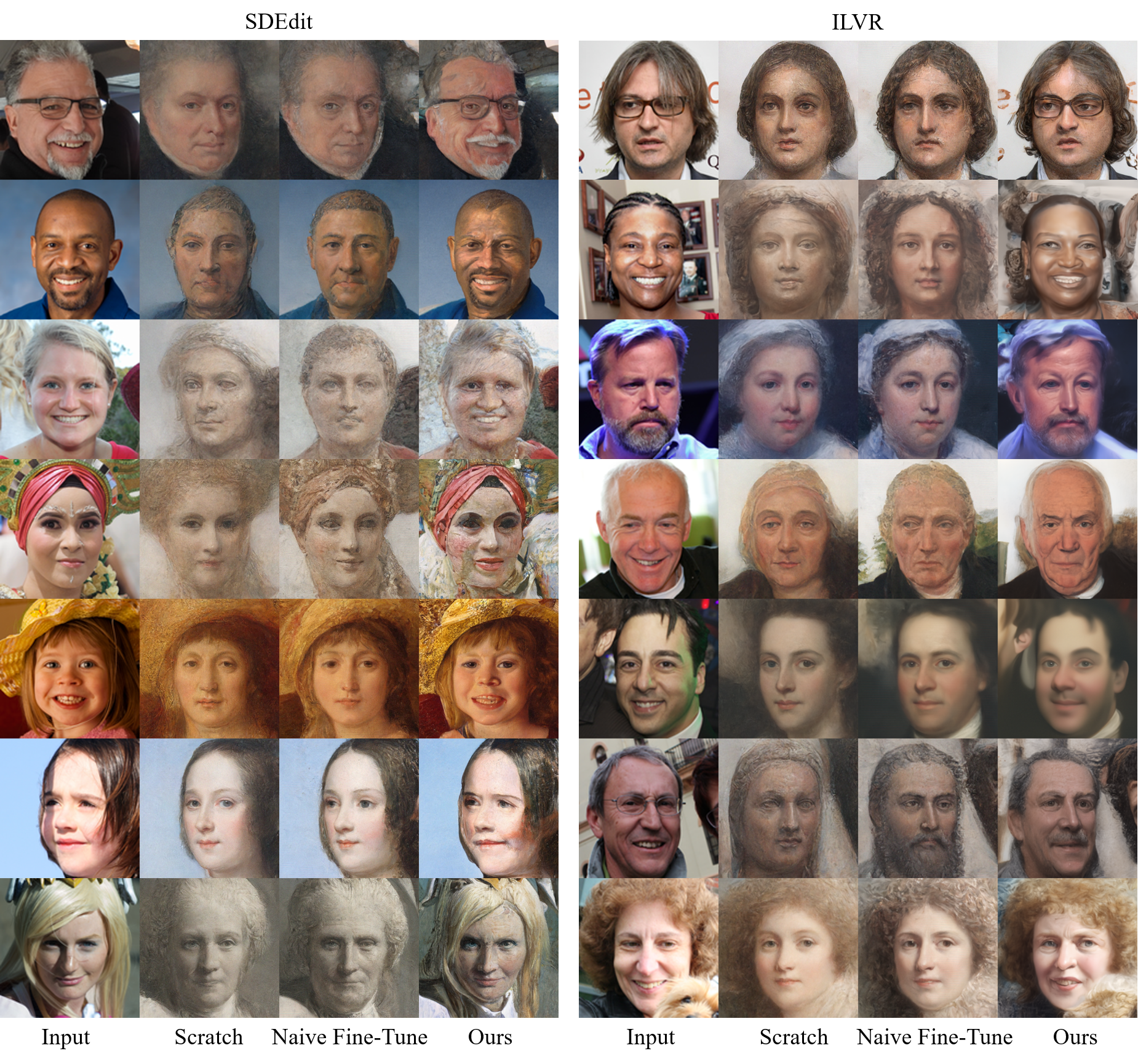}
\caption{Domain translation outputs using SDEdit and ILVR with fine-tuned diffusion models.}
\label{fig:sde_ilvr}
\end{figure*}

\begin{figure}[!ht]
    \centering
    \includegraphics[width=\linewidth]{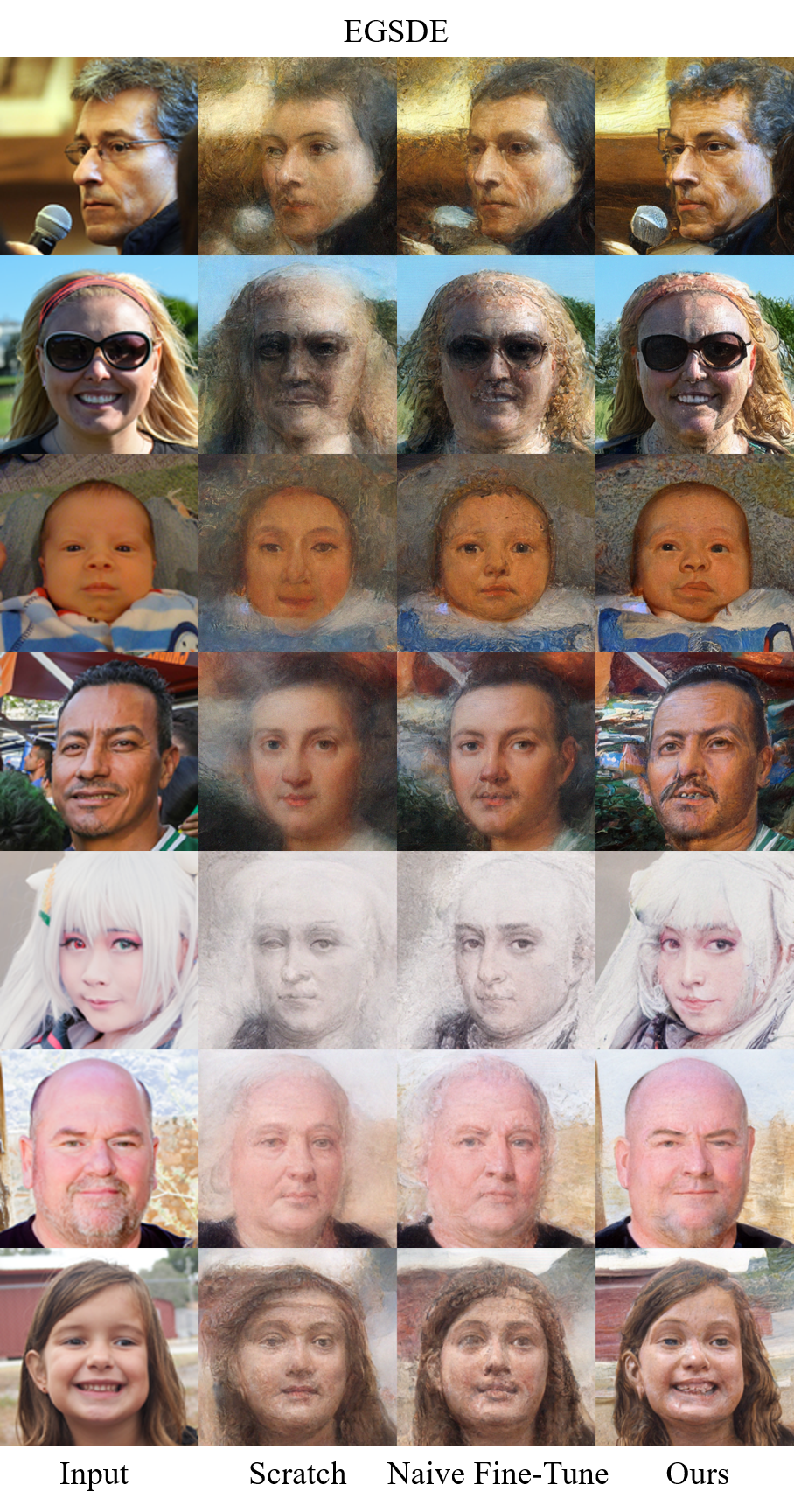}
\caption{Domain translation outputs using EGSDE with fine-tuned diffusion models.}
\label{fig:egsde}
\end{figure}

For all domain translation methods~\cite{meng2021sdedit,zhao2022egsde}, rather than editing in all diffusion time steps, authors used partial editing time steps for the best trade-off between \textit{realistic} and \textit{faithful}. 
Note that successful domain translation should be \textit{realistic} to fit the style of the target domain and \textit{faithful} to ensure that the various attributes from the input image are accurately preserved.
The larger the editing time steps, the more \textit{realistic} the translated outputs, but the less \textit{faithful} it becomes, losing the crucial attributes from the source images.
As a result, various domain translation methods generally adopt an editing range of $0.5T$ in their experiments for the best trade-off between \textit{realism} and \textit{faithfulness}. Following this, in the main manuscript, we adopt an editing range of $0.5T$ for all experiments on MetFaces.
For AAHQ experiments, we found that an editing range of $0.5T$ is less \textit{realistic}, thus we increase the editing range into $0.625T$ for all experiments on AAHQ.  
For the EGSDE~\cite{zhao2022egsde}, we train domain classifier for domain-independent energy function using 10K FFHQ~\cite{karras2019style} samples and each training datasets (entire MetFaces dataset or limited AAHQ dataset used for training of diffusion models). We used the official training code provided by the author, with 1500 and 4500 training iterations, respectively. 
For the domain-specific energy function, we use a downsampler with a downsampling factor of 32.
For the comprehensive comparison, we provide more results of domain translation in \cref{fig:sde_ilvr,fig:egsde}.

\section{Experimental Details in Text-Guided Image Manipulation}
\begin{figure*}[!ht]
    \centering
    \includegraphics[width=\textwidth]{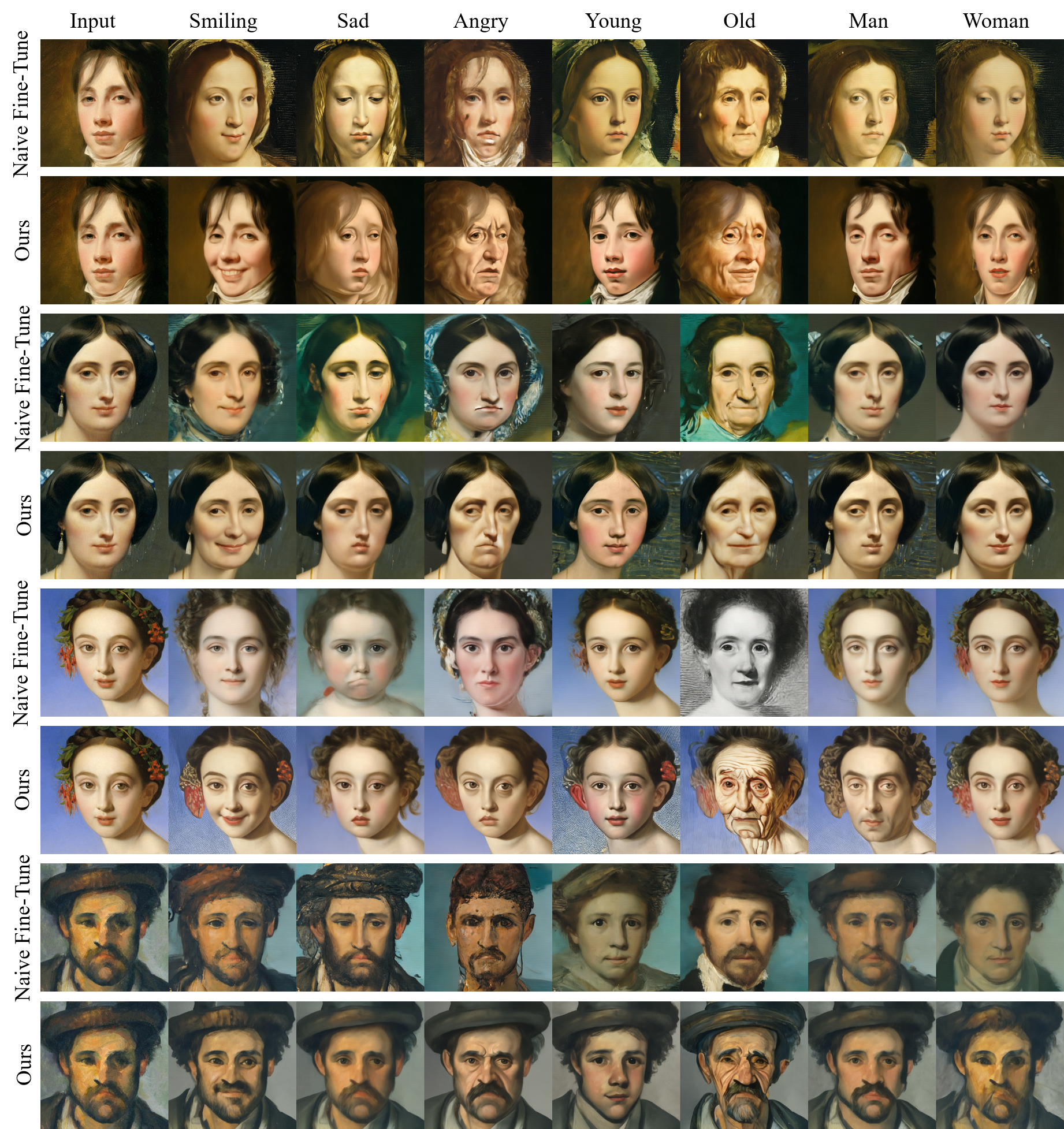}
\caption{Various results of text-guided image manipulation, where the input is close to the human face.}
\label{fig:supp_text_guided1}
\end{figure*}

\begin{figure*}[!ht]
    \centering
    \includegraphics[width=\textwidth]{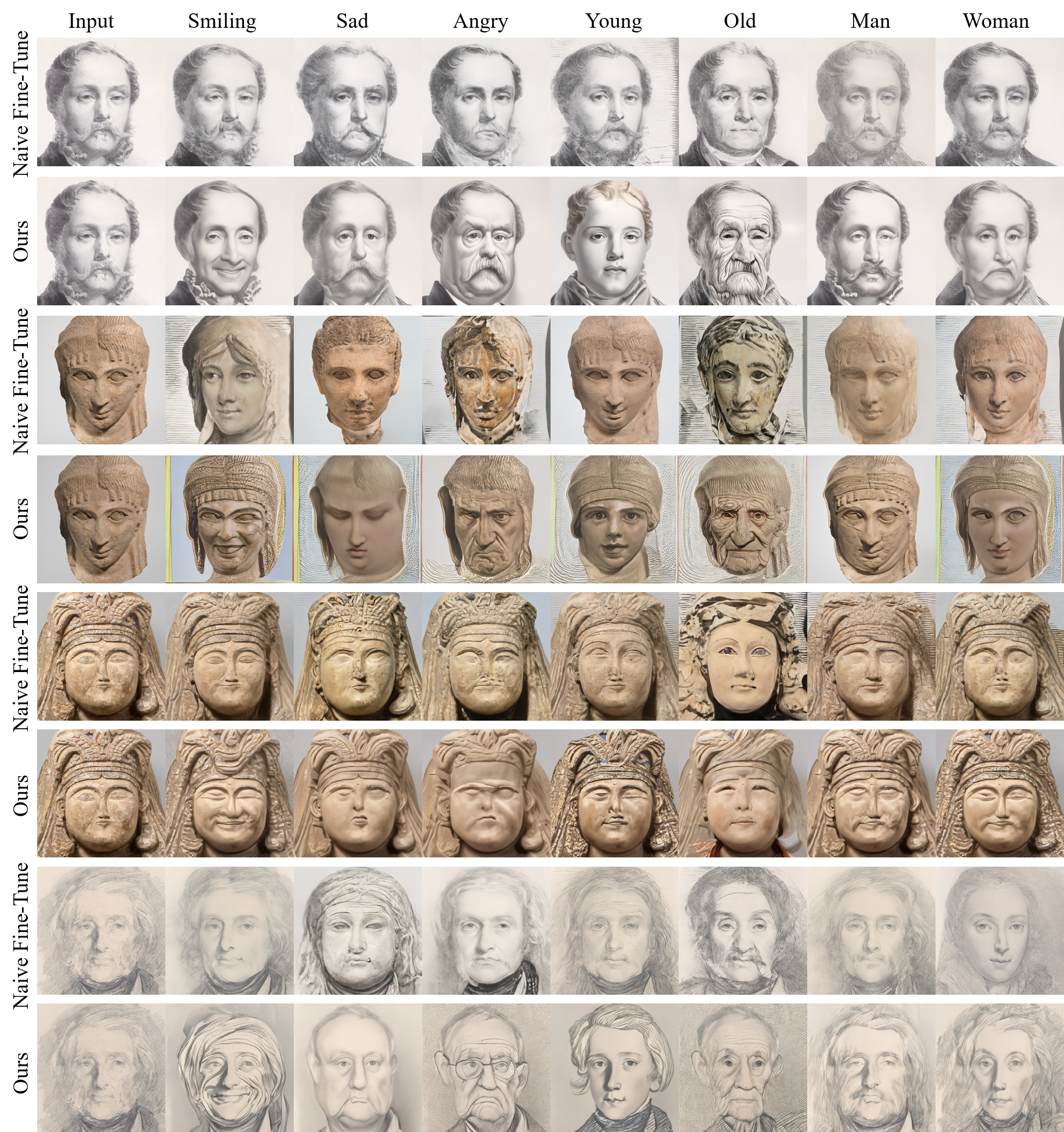}
\caption{Various results of text-guided image manipulation, where the input is not relatively close to the human face.}
\label{fig:supp_text_guided2}
\end{figure*}

In the main manuscript, we utilize Asyrp~\cite{kwon2022diffusion} for the text-guided image manipulation method.
We use officially provided training and inference code to implement Asyrp and use training step 50 and inversion step 100 in all experiments.
We utilize 100 training images from the MetFaces and AAHQ for Asyrp and the training epoch is set to 5. 
However, we found that in some facial expressions such as \textit{sad, angry}, and \textit{old}, training epoch 5 results in a bad bias that the manipulated faces get too old. As a result, we set the training epoch as 3 for these facial expressions.
The editing range is set to $0.5T$. 
For quantitative evaluation, we use 500 and 400 test images for MetFaces and AAHQ, respectively, and use 5 scripts (\textit{smiling, sad, angry, young and old}).
For the comprehensive comparison, we provide more results of text-guided translation in \cref{fig:supp_text_guided1,fig:supp_text_guided2} using text guidance \textit{smiling, sad, angry, young, old, man}, and \textit{woman}.
Overall, ours(SDFT) outperforms for the various expressions, even though the input image is not close to the real human face, while successfully preserving the identities.
We note that the provided results are not included in the training datasets for Asyrp. 

\section{More Results on Unconditional Image Generation}
\begin{figure}[!ht]
    \centering
    \includegraphics[width=\linewidth]{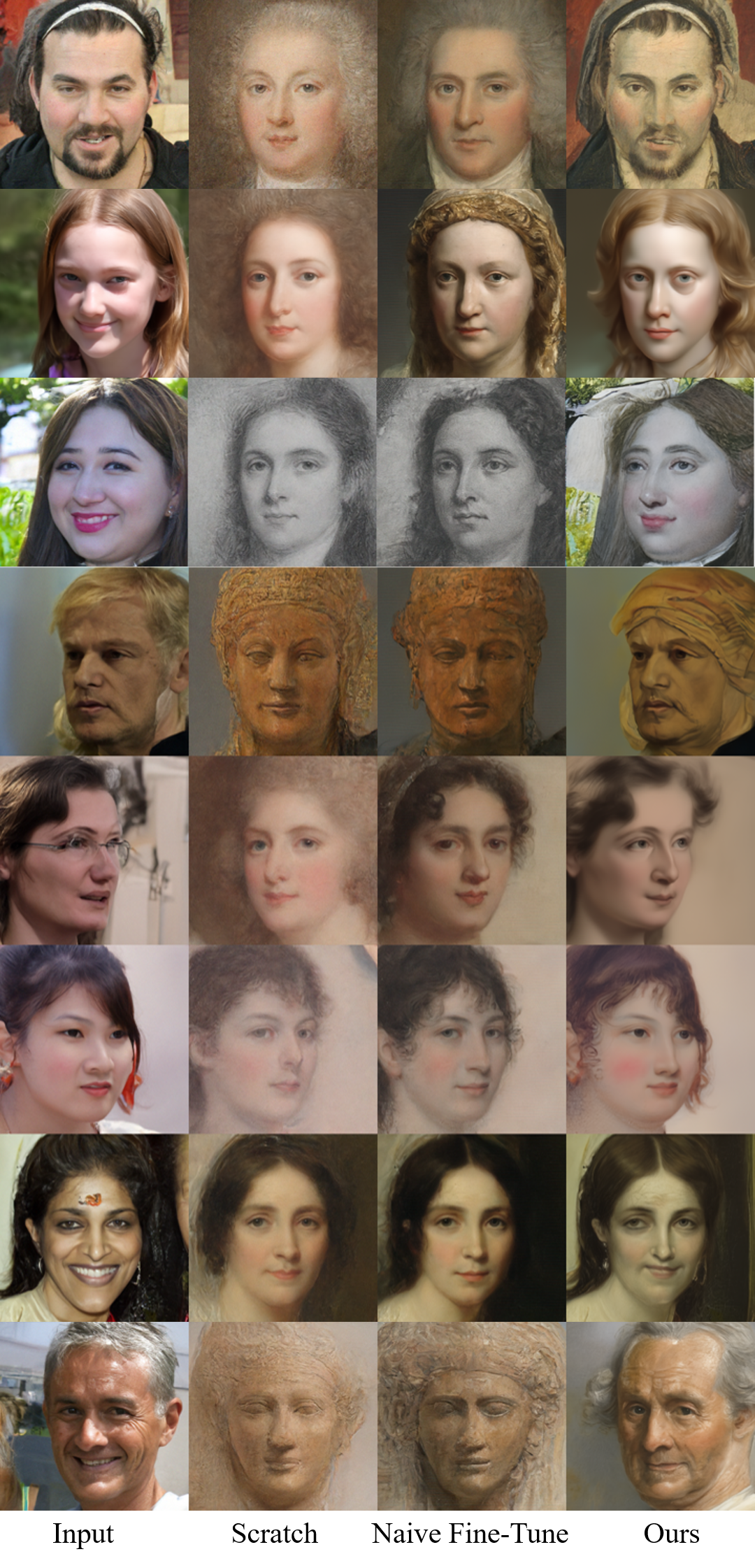}
\caption{Various results of unconditional image generation. Images in each row are generated from the same initial noise.}
\label{fig:supp_uncond}
\end{figure}

In \cref{fig:supp_uncond}, we provide more samples from unconditional image generation for the comprehensive comparison. As described in \cref{sec:supp_ddib}, the images generated from the same initial noise have aligned semantics. Since SDFT can preserve more diverse information from the source model, SDFT can generate more semantically aligned, diverse images from unconditional image generation.

